%% file: main.tex
\pdfoutput=1
\documentclass[11pt]{article}

\usepackage{acl}

\newif\ifanonymoussubmissionmode
\newif\ifarxivmode

\arxivmodefalse

\input{00_packages}
\input{01_commands}

\usepackage[T1]{fontenc}

\usepackage[utf8]{inputenc}

\usepackage{microtype}

\usepackage{inconsolata}

\usepackage{graphicx}

\title{Derivational Probing: Unveiling the Layer-wise Derivation \\ of Syntactic Structures in Neural Language Models}

\author{
    \textbf{Taiga Someya\textsuperscript{\scriptsize $\spadesuit$}}\,
    \textbf{Ryo Yoshida\textsuperscript{\scriptsize $\spadesuit$}}\,
    \textbf{Hitomi Yanaka\textsuperscript{\scriptsize $\spadesuit$$\heartsuit$}}
    \textbf{Yohei Oseki\textsuperscript{\scriptsize $\spadesuit$}}
    \\
    \textsuperscript{\scriptsize $\spadesuit$}The University of Tokyo\,
    \textsuperscript{\scriptsize $\heartsuit$}RIKEN
    \\
        \texttt{\{taiga98-0809,yoshiryo0617,hyanaka,oseki\}@g.ecc.u-tokyo.ac.jp}
}

\begin{document}
\maketitle
\begin{abstract}
Recent work has demonstrated that neural language models encode syntactic structures in their internal \emph{representations}, yet the \emph{derivations} by which these structures are constructed across layers remain poorly understood.
In this paper, we propose \emph{Derivational Probing} to investigate how micro-syntactic structures (e.g., subject noun phrases) and macro-syntactic structures (e.g., the relationship between the root verbs and their direct dependents) are constructed as word embeddings propagate upward across layers.
Our experiments on BERT reveal a clear bottom-up derivation: micro-syntactic structures emerge in lower layers and are gradually integrated into a coherent macro-syntactic structure in higher layers.
Furthermore, a targeted evaluation on subject-verb number agreement shows that the timing of constructing macro-syntactic structures is critical for downstream performance, suggesting an optimal timing for integrating global syntactic information.
\begin{center}
    \faGithub~\url{https://github.com/osekilab/derivational-probing}
\end{center}
\end{abstract}

\section{Introduction}

\begin{figure*}[t]
    \centering
    \includegraphics[width=\linewidth]{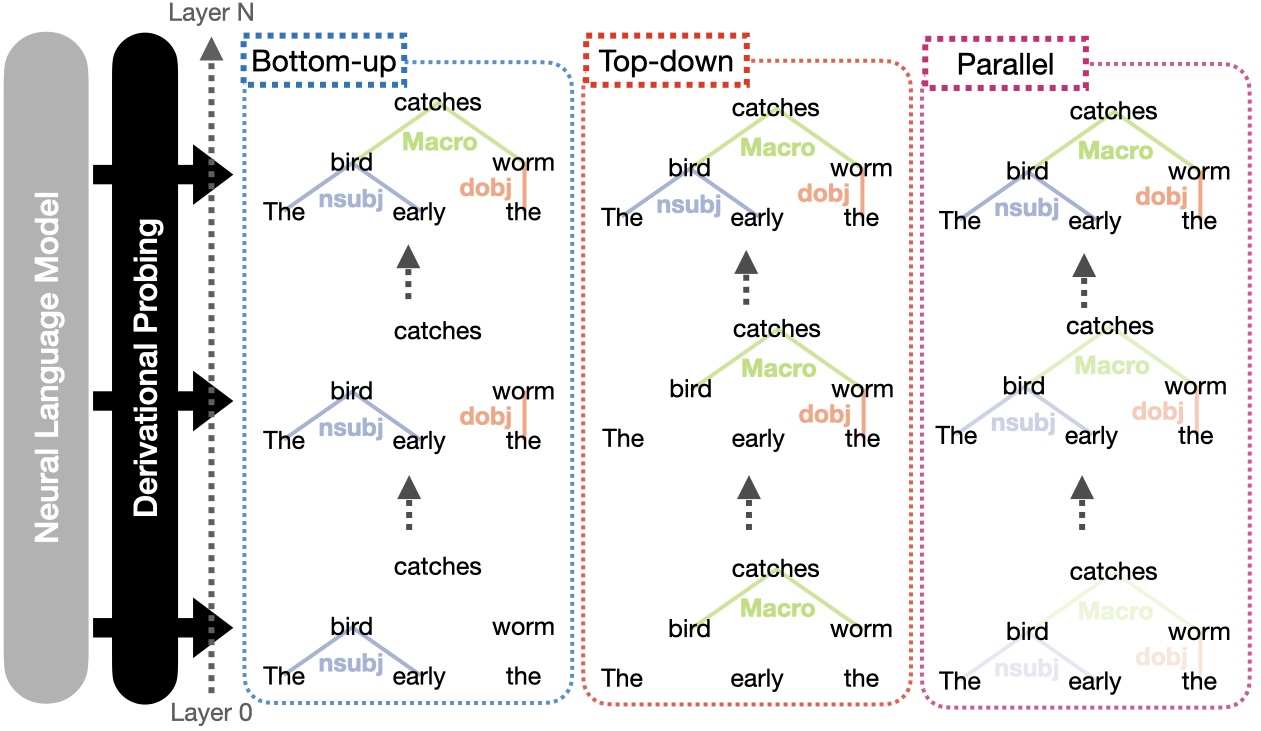}
    \caption{Derivational Probing investigates how syntactic structures are constructed across layers in neural language models. We illustrate three hypotheses for syntactic derivation.
    \textbf{Bottom-up:} Micro-syntactic structures, such as subject noun phrases or prepositional phrases, emerge in lower layers, and the macro-syntactic structure is formed in higher layers.
    \textbf{Top-down:} The macro-syntactic structure is formed in lower layers, with micro-syntactic structures refined in higher layers.
    \textbf{Parallel:} Micro- and macro-syntactic structures emerge in parallel.}
    \label{fig:derivation_probing}
    \vspace{-0.2cm}
\end{figure*}

Neural language models have achieved remarkable success across a wide range of natural language processing tasks.
However, significant uncertainty remains regarding what these models truly learn and how they represent linguistic knowledge.
This has spurred extensive research aimed at probing the linguistic capabilities of neural language models~\citep{10.1145/3639372,10.1162/coli_a_00492}.

A prominent line of inquiry is \emph{structural probing}, which directly analyzes word embeddings to uncover latent syntactic structures.
For example, \citet{hewitt2019structural} demonstrated that the geometric organization of the word embedding space in BERT~\citep{devlin-etal-2019-bert} encodes syntactic distances defined over dependency parse trees, providing evidence that the model captures syntactic information.
Yet, such work typically focuses on the static representations of the whole syntactic structures rather than the dynamic derivations by which these syntactic structures are built across layers.
Understanding not just the resulting representations but also how they are built across layers is essential for a more comprehensive understanding and could also lead to better insights into how these representations are used.

Meanwhile, \citet{tenney2019bert} introduced the \emph{expected layer} metric and investigated how different layers in BERT encode different types of linguistic information (e.g., part-of-speech tagging, syntactic parsing, semantic role labeling, and coreference resolution), revealing that the model encodes linguistic abstractions in a manner reflecting a traditional NLP pipeline.
However, their approach primarily relied on coarse-grained task accuracy measures, capturing only the overall effectiveness of each layer rather than examining the detailed, layer-wise construction of specific syntactic structures.
Consequently, how the syntactic structures are built across layers remains under-explored.

In this paper, we fill this gap by proposing \emph{Derivational Probing}---a method that integrates structural probing with the expected layer metric to probe derivation processes of syntactic structures in neural language models (\cref{fig:derivation_probing}). Our proposed method allows us to investigate how \emph{micro}-syntactic structures (e.g., subject and object noun phrases, prepositional phrases) and \emph{macro}-syntactic structures (e.g., the relationship between the root verbs and their direct dependents) are constructed across layers.

Applying Derivational Probing to BERT~\citep{devlin-etal-2019-bert}, our experiments reveal a clear bottom-up derivation, in which micro-syntactic structures emerge in lower layers and are gradually integrated into a coherent macro-syntactic structure in higher layers. Furthermore, our targeted analysis on a subject-verb number agreement task shows that even when the final syntactic structure is correct, the specific layers at which the macro-syntactic structure is constructed significantly affect downstream performance. This suggests the existence of an optimal timing for integrating global syntactic information.

Overall, our findings offer new insights into the internal mechanisms by which neural language models construct syntactic structures and underscore the importance of examining derivation processes across layers to improve the interpretability of neural language models.\looseness-1

\section{Related Work}
Attention-based analyses \citep[e.g.,][]{clark-etal-2019-bert,vig-belinkov-2019-analyzing} have demonstrated that certain transformer heads tend to align with dependency relations, providing evidence that Transformer language models capture linguistic dependency relations in their attention weights.\looseness-1

In contrast, \citet{hewitt2019structural} introduced a structural probe with a linear transformation from hidden representations into a space where Euclidean distances reflect dependency tree distances.
This approach revealed that full syntactic trees are implicitly encoded in models such as BERT.
Building on this, later work refined the approach by incorporating non-linear mappings \citep[e.g.,][]{white-etal-2021-non}, enforcing constraints such as orthogonality \citep{limisiewicz-marecek-2021-introducing}, and using a controlled corpus to isolate the effect of syntax \citep{hall-maudslay-cotterell-2021-syntactic}.\looseness-1

Other studies have refined structural probing by quantifying context-dependent syntactic signals in deeper layers---for example, conditional probing \citep{hewitt-etal-2021-conditional} and information gain metrics \citep{kunz-kuhlmann-2022-linguistic}---but these methods focus on the performance of specific probing tasks (e.g., POS-tagging) rather than where the syntactic structures are constructed.\looseness-1

In contrast, our proposed method specifically tracks how each subgraph in the syntactic tree develops as information propagates through the network layers.
By analyzing the evolution of individual syntactic components---from micro-syntactic structures to the assembly of the macro-syntactic structure---we offer a more granular perspective on the incremental construction of syntax, complementing and extending previous layer-wise analyses.

\section{Technical Preliminaries}
In this section, we review foundational methods from prior research: structural probing to assess the presence and quality of syntactic representations and the expected layer metric for quantifying how linguistic information gradually builds up across successive layers within language models.

\subsection{Structural Probing} \label{sec:structural-probe}
\newcite{hewitt2019structural} introduced the \emph{structural probe} as a method to evaluate whether contextual word representations encode syntactic information.
Given a sentence \(\mathbf{s} = w_1 \cdots w_t\), each token is represented by a $d$-dimentional contextual embedding \(\mathbf{h}_i \in \mathbb{R}^{d}\) (e.g., the output embedding of a model like BERT).
The goal of the structural probe is to find a linear transformation that maps these embeddings to a space where the Euclidean distances approximate the true syntactic distances between words.

Specifically, for any two words \(w_i\) and \(w_j\) in a sentence, we define the transformed distance as:
\begin{equation} \label{eq:structural_distance}
    d_\mathbf{B}(\mathbf{h}_i, \mathbf{h}_j) = \| \mathbf{B}\mathbf{h}_i - \mathbf{B}\mathbf{h}_j \|_2,
\end{equation}
where \(\mathbf{B} \in \mathbb{R}^{d' \times d}\) is a learnable projection matrix.
The true syntactic distance, \(\Delta_{ij}\), is typically defined as the number of edges on the shortest path between \(w_i\) and \(w_j\) in the dependency parse tree of the sentence. The probe is trained by minimizing an objective that penalizes the discrepancy between the predicted distances and \(\Delta_{ij}\):
\begin{equation}\label{eq:structural_loss}
    \mathcal{L} = \frac{1}{|\mathbf{s}|^2} \sum_{i=1}^{|\mathbf{s}|} \sum_{j=i+1}^{|\mathbf{s}|} \left| \Delta_{ij} - d_\mathbf{B}(\mathbf{h}_i, \mathbf{h}_j) \right|.
\end{equation}

This formulation encourages the linear transformation \(\mathbf{B}\) to capture the syntactic structure encoded in the contextual representations, enabling the recovery of parse trees via \citeposs{prim_1957} algorithm, a greedy algorithm that constructs minimum spanning trees by iteratively adding the lowest-weight edge connecting a new node to the growing tree.

\subsection{Expected Layer} \label{sec:expected-layer}
The \emph{expected layer} metric introduced by \newcite{tenney2019bert} was initially developed to identify the layers within BERT responsible for solving various linguistic tasks.
Specifically, the metric was used to capture at which layers broad linguistic abilities (e.g., part-of-speech tagging, syntactic parsing, semantic role labeling) emerge.
\newcommand{\ScalerMixedEmbeddings}{\bm{m}}
\citeauthor{tenney2019bert} defined \emph{scaler-mixed embeddings} $\ScalerMixedEmbeddings_i^{\ell} \in \mathbb{R}^{d}$ as the weighted average of embeddings from the bottom layer up to layer $\ell$:
\begin{equation}\label{eq:scalar-mixed-embedding}
    \ScalerMixedEmbeddings_i^{\ell} = \gamma \sum_{k=0}^\ell \mixingweight_{k}\,\mathbf{h}_i^{k},
\end{equation}
where $\mixingweight = \softmax(\mathbf{a})\,(\mathbf{a} \in \mathbb{R}^{\ell+1})$  is learnable scalar mixing weights and \(\gamma\) is a learnable scaling factor, following \citet{peters-etal-2018-deep}.

By measuring performance at layer index $\ell$, denoted by $S(\ell)$, and tracking its improvements across layers, \citeauthor{tenney2019bert} defined the expected layer  to reflect the layer at which the relevant linguistic task information is predominantly captured:
\begin{equation}\label{eq:expected-layer}
    E[\ell] = \sum_{\ell=1}^L \frac{S(\ell) - S(\ell-1)}{\sum_{\ell=1}^L (S(\ell) - S(\ell-1))} \ell.
\end{equation}
This is the weighted average of layer indices, where each layer's weight corresponds to its relative contribution to the overall performance improvement.
It was initially proposed to broadly characterize the hierarchical progression of different linguistic capabilities within transformer models, rather than pinpointing the exact layers at which specific syntactic structures are built.

\section{Derivational Probing} \label{sec:derivation_probing}

Building upon these prior techniques, we propose \emph{Derivational Probing}, a novel method explicitly designed to investigate the dynamic construction of syntactic structures across the layers of neural language models.

Our approach effectively combines expected layer metric \citep{tenney2019bert} with the structural probing~\citep{hewitt2019structural}, enabling a detailed analysis of how syntactic information accumulates across model layers.
Specifically, for each layer $\ell$, we use scalar-mixed embeddings as defined in \cref{eq:scalar-mixed-embedding} and compute pairwise distances:
\begin{equation}\label{eq:derivation-distance-new}
    d_\mathbf{B_\ell}(\ScalerMixedEmbeddings_i^{\ell}, \ScalerMixedEmbeddings_j^{\ell}) = \|\mathbf{B_\ell}\ScalerMixedEmbeddings_i^{\ell} - \mathbf{B_\ell}\ScalerMixedEmbeddings_j^{\ell}\|_2.
\end{equation}
We then train the transformation matrix \(\mathbf{B_k}\) to minimize discrepancies with true dependency parse distances, analogous to structural probing.

This integration allows us to calculate the expected layer for each syntactic subgraph (micro- and macro-syntactic structures defined in detail later) and perform a fine-grained, quantitative analysis of their construction across model layers.
We use the Unlabeled Undirected Attachment Score (UUAS) for each layer $\ell$ as $S(\ell)$, defined as the proportion of correctly predicted edges to the total number of edges in the reference dependency parse, without considering edge labels or direction.

\begin{figure}[h]
    \centering
    \includegraphics[width=\linewidth]{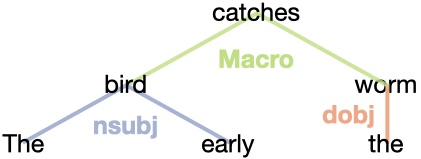}
    \caption{\emph{Macro-syntactic structure} (\depMacro{}) and \emph{micro-syntactic structures} (\depnsubj{}\,and \depdobj{}). }
    \label{fig:structure}
\end{figure}

To better understand the derivation strategy that models employ when constructing a syntactic tree, we introduce a distinction between \emph{macro-syntactic structures} (the root verb with its direct dependents) and \emph{micro-syntactic structures} (local components, such as subordinate phrases like \depnsubj{}) (\cref{fig:structure}).
This distinction is motivated by our interest in whether models construct syntactic trees top-down, bottom-up, or in a parallel fashion.
To empirically evaluate which of these hypotheses is most plausible, we adopt the following methodological approach: For both micro-syntactic structures and macro-syntactic structures, we (1) construct the full parse tree using a minimum spanning tree algorithm, (2) extract the relevant edges (as highlighted in \cref{fig:structure}), and (3) compute the UUAS by comparing these edges to the reference parse. By tracking UUAS improvements across layers, we calculate the expected layer \(E[l]\) for each structure, revealing the layers at which different syntactic subgraphs are effectively constructed.

We next provide detailed descriptions of each hypothesis.
\paragraph{Bottom-up derivation.} A bottom-up derivation first constructs micro-syntactic structures and subsequently integrates these into macro-syntactic structures, ultimately forming a complete dependency tree.
We refer to this as a ``bottom-up derivation'' because it resembles the construction order of the arc-standard transition-based dependency parser \citep{nivre-2004-incrementality}.
Arc-standard parsing utilizes a stack-based transition system and constructs a dependency tree in a bottom-up manner: dependents must be fully processed and attached to their heads before those heads themselves are incorporated into macro-syntactic structures.
Under this hypothesis, models initially identify micro-syntactic structures---such as the internal phrase structures of subjects and objects---in lower layers, which are then progressively combined into a coherent macro-syntactic hierarchy at higher layers.

\paragraph{Top-down derivation.} A top-down derivation, in contrast, begins by establishing macro-syntactic structures and subsequently refines these by incorporating detailed micro-syntactic dependencies.
We term this approach a ``top-down derivation'' because its construction order aligns closely with the head-driven transition-based parser proposed by \citet{hayashi-etal-2012-head}.
Their algorithm explicitly predicts dependent nodes from head nodes, progressively building syntactic structures from head to dependent, thus genuinely following a top-down, predictive parsing order.
Under this hypothesis, models prioritize the recognition of macro-syntactic structures before refining micro-syntactic struvtures.

\paragraph{Parallel derivation.} Finally, an alternative hypothesis is that models construct micro- and macro-syntactic structures concurrently, with local dependencies and the global structure forming at roughly the same rate across layers. This hypothesis is less clearly aligned with traditional dependency parsing algorithms, as most classical approaches tend to favor either bottom-up or top-down derivations.

\paragraph{Notes on the term ``derivation''.} Here, we explicitly use the term \emph{derivation} (strategy) throughout this paper rather than ``parsing strategy'' to clearly distinguish two related but distinct concepts.
While ``parsing strategy'' generally refers to methodological choices for \emph{incrementally} constructing a parse tree (such as bottom-up or top-down), our use of ``derivation'' specifically captures an \emph{atemporal} process describing how syntactic structures progressively emerge across the internal layers of a language model given the full sentence context, emphasizing \emph{layer-wise} structural development rather than sequential, \emph{left-to-right} incremental processing.

\section{Experimental Setup} \label{sec:experimental_setup}
\subsection{Data} \label{sec:data}

\begin{table*}[t]
    \centering
        \begin{tabular}{@{}ll@{}}
        \toprule
        \textbf{Structure Set} & \textbf{Example Sentence} \\
        \midrule
        \depMacro{}, \depnsubj{}, \depdobj{} & The concert caused a major stir. \\
        \depMacro{}, \depnsubj{}, \depprep{} & The match ended in a goalless draw. \\
        \depMacro{}, \depnsubj{}, \depattr{} & Her parents were music professors. \\
        \depMacro{}, \depnsubj{}, \depprep{}, \depdobj{} & The film received positive reviews from critics. \\
        \bottomrule
        \end{tabular}
    \caption{Example sentences for each primary structure set described in \Cref{sec:data}}
    \label{tab:example_structure_sets}
\end{table*}

We utilize the Wikitext-103 dataset~\citep{merity2016pointer} as our primary source of natural language, parsing each sentence with spaCy's dependency parser (\textsc{en\_core\_web\_lg})~\citep{Honnibal_spaCy_Industrial-strength_Natural_2020}.
To focus on the language model's ability to construct syntactic structures in a clear-cut setting, we restrict our analysis to single-clause sentences by excluding those with relative clauses or clausal subjects.
Additionally, we filter out sentences containing dependency relations such as ``dep'' (unclassified dependents) and punctuation marks other than sentence-final punctuation to minimize noise.

Following the definitions introduced in the previous section (\cref{sec:derivation_probing}), we group sentences based on dependency relations emanating from the root verb, thereby distinguishing between the overall (macro-syntactic; \depMacro{}) structure and subordinate (micro-syntactic) structures.
We retain only those groups that represent more than 10\% of the data, focusing our analysis on the predominant structure sets.
This filtering results in four primary structure sets (See \cref{tab:example_structure_sets} for examples): (1) \depMacro{} with micro relations \depnsubj{} and \depdobj{}; (2) \depMacro{} with micro relations \depnsubj{} and \depprep{}; (3) \depMacro{} with micro relations \depnsubj{} and \depattr{}; and (4) \depMacro{} with micro relations \depnsubj{}, \depprep{}, and \depdobj{}.

From the resulting dataset, we randomly sample 50,000 sentences, partitioning them into 40,000 for training, 5,000 for validation, and 5,000 for testing.

\subsection{Models}
We employ two pre-trained language models: BERT-base\footnote{\url{https://huggingface.co/google-bert/bert-base-cased}} and BERT-large\footnote{\url{https://huggingface.co/google-bert/bert-large-cased}} (cased)~\citep{devlin-etal-2019-bert}.
BERT-base uses 12 layers, 12 heads, and a 768-dimensional hidden state, while BERT-large uses 24 layers, 16 heads, and a 1024-dimensional hidden state. These models provide a range of capacities, allowing us to investigate differences in how syntactic structures are constructed across models.

We focus specifically on BERT because our method is designed to examine the \emph{atemporal}, layer-wise derivation of syntactic structures given entire sentences.
In contrast, autoregressive language models such as GPT-2 process information incrementally in a left-to-right manner, and the \emph{temporal}, token-wise derivation of syntactic structures cannot be probed via our method~\citep[cf.][]{eisape-etal-2022-probing}. Nevertheless, our method is still applicable to the word embeddings of autoregressive language models such as GPT-2, and we report GPT-2 results in \cref{sec:gpt2_result}.

For each model, we probe all layers to determine the progression of syntactic information and compute the expected layer at which specific structures emerge.
We conduct training with five different random seeds and report the average performance along with the standard deviation.
Additional hyperparameters and training details are provided in \cref{sec:hparams}.

\section{Results}
\subsection{Overall UUAS Performance}
\begin{figure}[t]
    \centering
    \includegraphics[width=\linewidth]{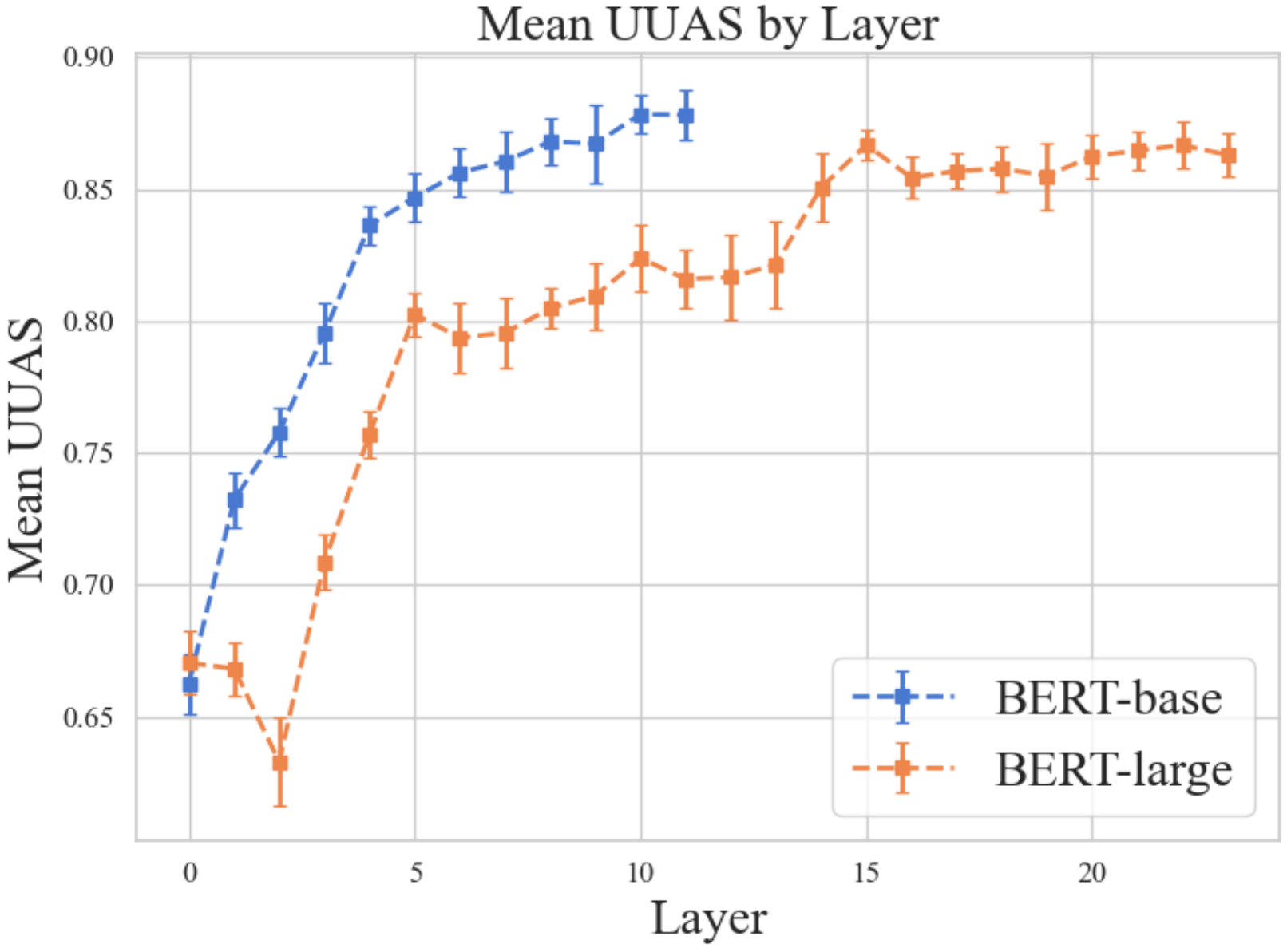}
    \caption{Global UUAS by each layer for each model. Error bars represent standard deviations across 5 random seeds.}
    \label{fig:global_uuas}
\end{figure}

\begin{figure*}[t]
    \centering
    \includegraphics[width=\linewidth]{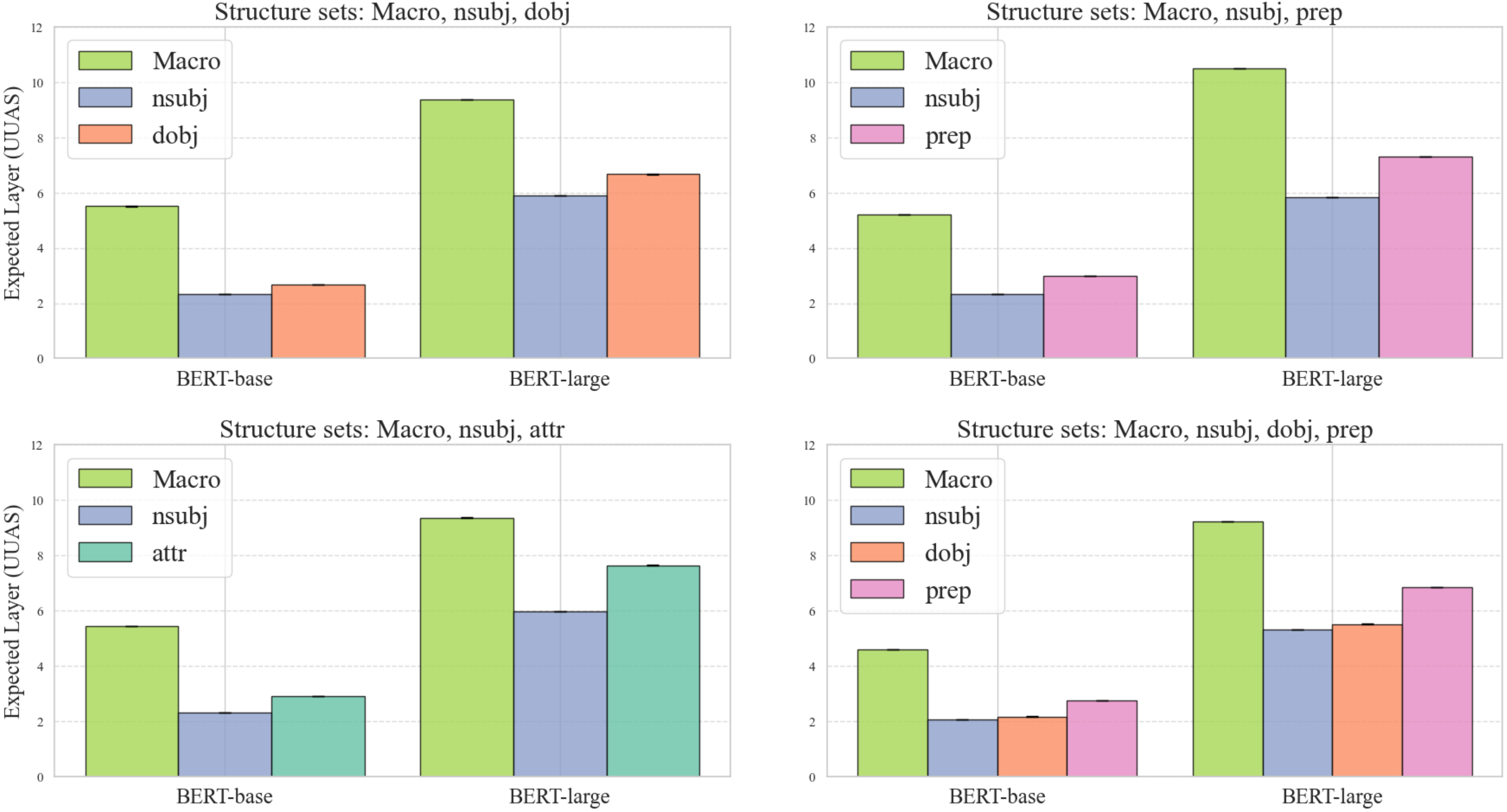}
    \caption{Expected layer for each model across different structure sets. Error bars represent standard deviation across 5 random seeds.}
    \label{fig:result_by_structure_sets}
\end{figure*}

As a sanity check to verify whether our models exhibit overall trends similar to those reported in previous studies, we conducted an experiment measuring the test set UUAS for overall sentence structures across layers for each model (\Cref{fig:global_uuas}).
BERT-base and BERT-large display similar trends, with the UUAS score saturating around the middle layers. BERT-large shows slightly slower improvement, likely reflecting its deeper architecture and larger capacity.
These trends mostly align with previous findings~\citep{hewitt2019structural} that neural language models tend to exhibit peak UUAS performance in their middle layers.
However, unlike previous studies, we do not observe a decrease in average UUAS in later layers, which we attribute to our method of computing word embeddings as a weighted average from layer 0 to layer $\ell$ (\cref{eq:scalar-mixed-embedding}).

\subsection{Expected Layer Across Structure Sets}
\begin{figure*}[t]
    \centering
    \includegraphics[width=\linewidth]{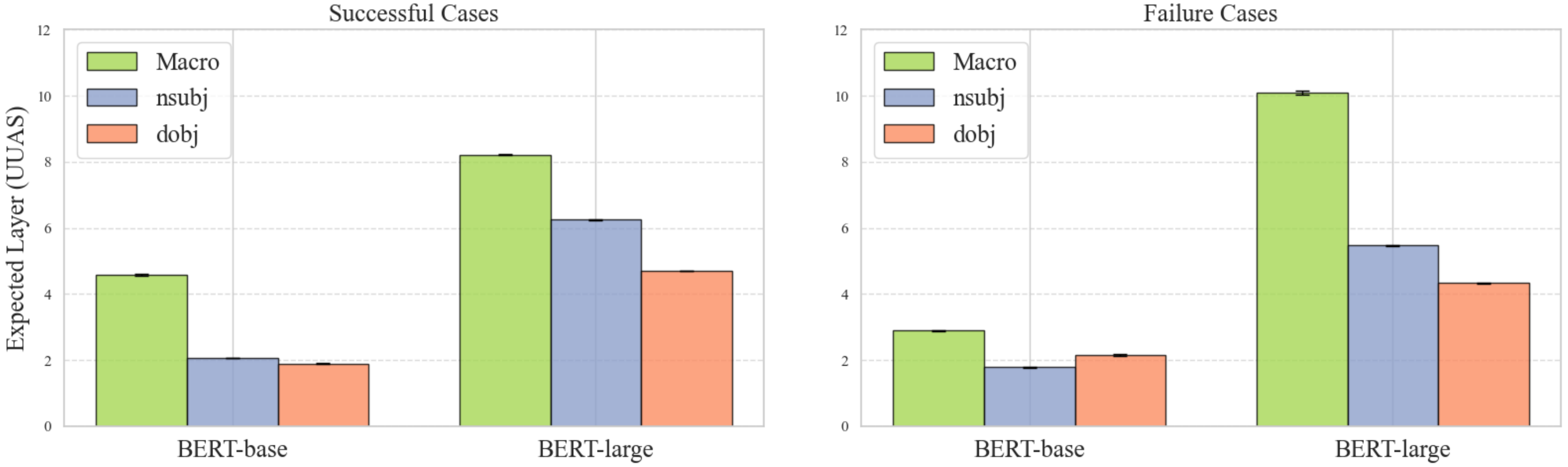}
    \caption{Expected layers for syntactic structures in successful and failed subject-verb agreement cases. Error bars show standard deviations across 5 random seeds.}
    \label{fig:tse_result}
\end{figure*}

\begin{figure*}[t]
    \centering
    \includegraphics[width=\linewidth]{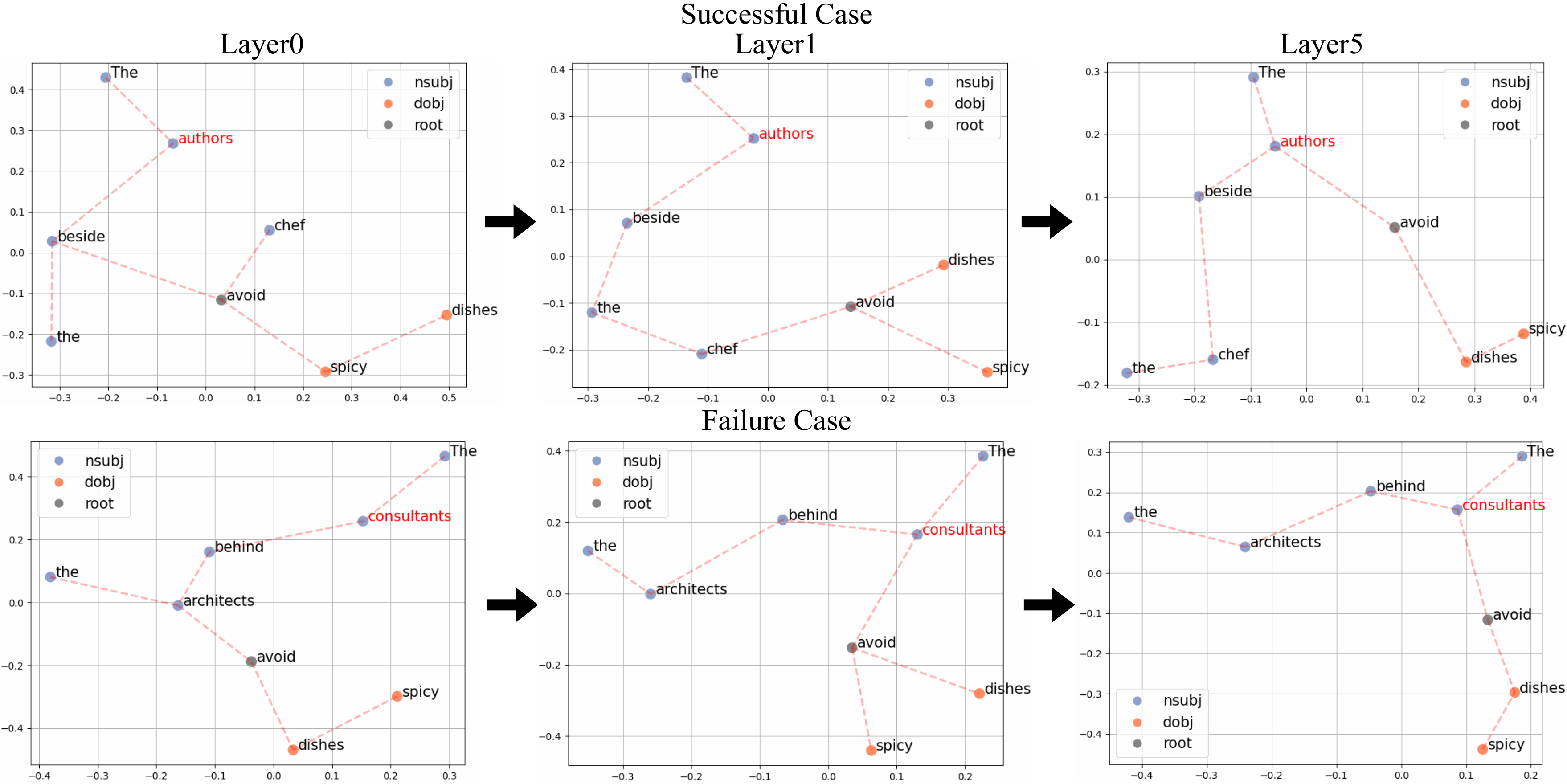}
    \caption{Derivation process visualizations for BERT-base on subject-verb agreement for a successful case (``The authors beside the chef avoid spicy dishes.'') and a failure case (``The consultants behind the architects avoid spicy dishes.''). Red highlights indicate the correct subject.}
    \label{fig:derivation_bert_base}
\end{figure*}

\begin{figure*}[t]
    \centering
    \includegraphics[width=\linewidth]{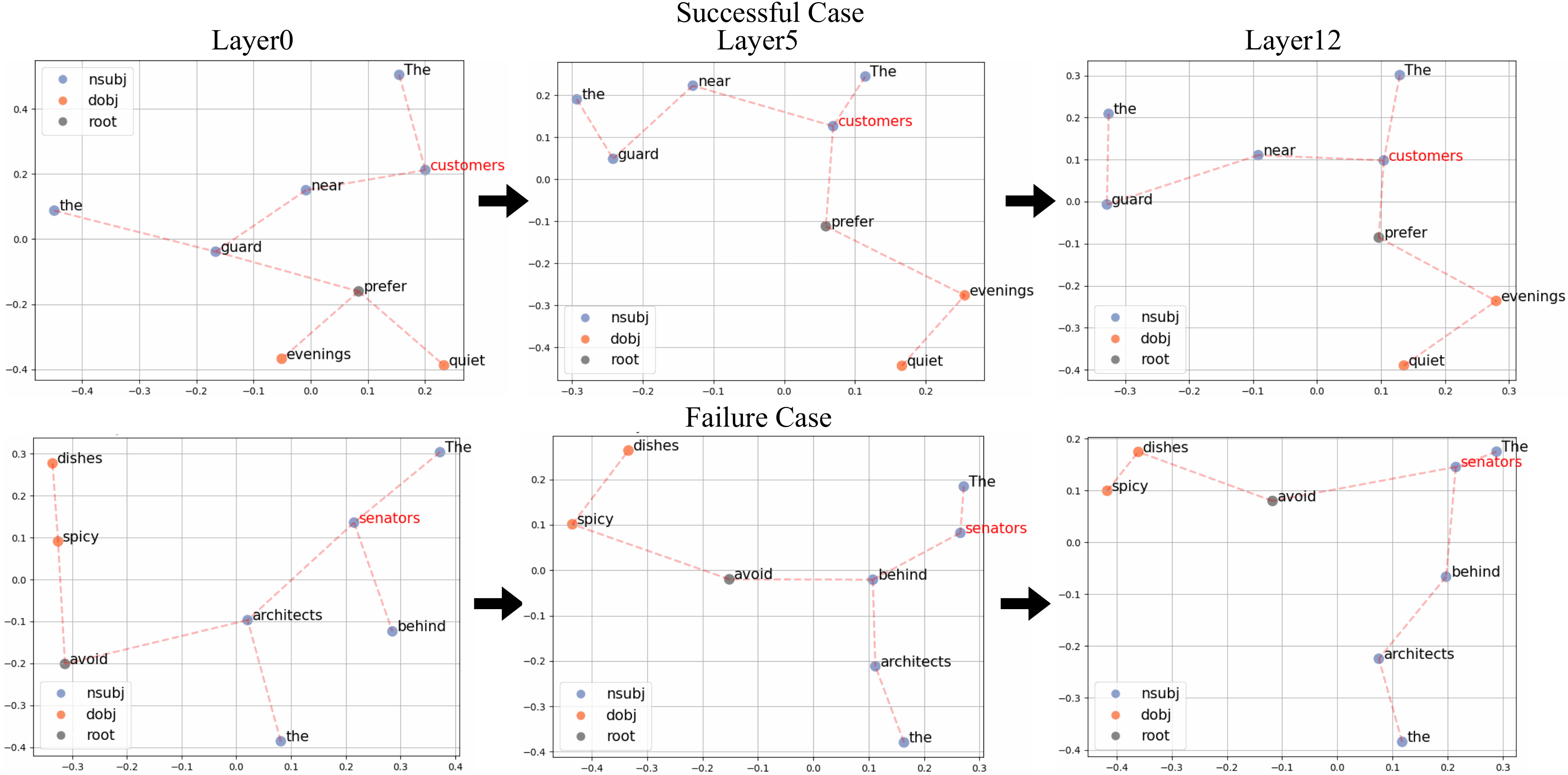}
    \caption{MDS visualizations of syntactic structure evolution in BERT-large for subject-verb agreement for a successful case (``The customers near the guard prefer quiet evenings.'') and a failure case (``The senators behind the architects avoid spicy dishes.''). Red highlights indicate the correct subject.}
    \label{fig:derivation_bert_large}
\end{figure*}

\Cref{fig:result_by_structure_sets} summarizes the expected layers for each syntactic structure within the four primary structure sets (\cref{sec:data}), for both BERT-base and BERT-large.

In both BERT-base and BERT-large, the macro-syntactic structure consistently exhibits the highest expected layer across all sets, whereas micro-syntactic structures such as \depnsubj{}, \depdobj{}, and \depprep{} tend to appear in lower layers.
This suggests a \emph{bottom-up} derivation process in which micro-syntactic structures (e.g., subject or object phrases) are constructed earlier, and these components are gradually integrated into a coherent macro-syntactic structure in later layers.
This observation is consistent with prior work on BERT, which shows that local information (e.g., POS tags) is captured early, while more abstract global structures emerge later \citep[cf.][]{tenney2019bert}.
Notably, this pattern holds for both BERT-base and BERT-large, although the overall expected layers are slightly higher in BERT-large—likely reflecting its deeper architecture and larger capacity.

\section{Detailed Analysis: Subject-Verb Agreement Task}
\subsection{Experimental Setup}
To investigate how the process and layers involved in syntactic structure construction relate to model performance, we conduct a detailed analysis on subject-verb agreement using sentences with intervening nouns (``attractors''), following the approach of \citet{marvin-linzen-2018-targeted} with some modification.
We sampled 1,000 positive (grammatical) and 1,000 negative (ungrammatical) sentences.
All of the sampled sentences have a subject followed by a prepositional phrase, a verb, and a direct object noun phrase.
They thus are categorized into \depMacro{}, \depnsubj{}, \depdobj{} structure sets.
In our modification of their templates, each of \depMacro{}, \depnsubj{}, \depdobj{} is required to contain more than one word.
\a. The senators behind the brilliant architect \underline{avoid} spicy dishes.
\b. *The senators behind the brilliant architect \underline{avoids} spicy dishes.

This ensures that we can extract meaningful subgraphs within each syntactic substructure.

We first evaluate model performance on this task by computing pseudo-whole-sentence probabilities~\citep{salazar-etal-2020-masked}.
Specifically, we calculate the probability of each token by masking it one by one and then aggregate these token-level probabilities to derive an overall sentence probability.
We expect the model to assign higher pseudo-probabilities to grammatical sentences compared to ungrammatical ones.
We then analyze how the syntactic construction process differs between cases where the model performs well and those where it fails.\looseness-1

Furthermore, to visualize the evolution of syntactic structures across model layers, we employ Multidimensional Scaling (MDS).
Specifically, we apply scikit-learn's MDS implementation \citep{scikit-learn} with default parameters to word embeddings projected by our structural probe, allowing us to illustrate clearly how syntactic representations develop across different model layers.

\subsection{Results}
Overall, BERT-base correctly assigned higher pseudo-whole-sentence probabilities to grammatical sentences in 984 out of 1,000 examples, whereas BERT-large achieved correctness in 983 out of 1,000 cases.
Despite their similar overall accuracies, we observe distinct patterns between BERT-base and BERT-large (\cref{fig:tse_result}).\looseness-1

\paragraph{BERT-base.} BERT-base frequently failed when macro-syntactic structures were established prematurely, potentially restricting the incorporation of essential micro-syntactic details.
As illustrated in \cref{fig:derivation_bert_base}, successful cases show a sequential pattern where BERT-base first constructs micro-syntactic structures within the subject phrase in early layers, subsequently aligning the subject (\textit{authors}) with the verb (\textit{avoid}) around layer 5 after stabilizing the internal subject dependencies.
In contrast, failure cases reveal premature alignment of macro-syntactic structures, with the subject (\textit{consultants}) prematurely linked to the verb (\textit{avoid}) before fully establishing necessary micro-syntactic details.
This premature commitment might have negatively impacted the overall syntactic representation, disrupting correct subject-verb agreement.\looseness-1

\paragraph{BERT-large.} BERT-large exhibited higher expected layers for macro-syntactic structures in failure cases, suggesting delayed integration of macro-syntactic information.
\Cref{fig:derivation_bert_large} illustrates representative successful and unsuccessful cases for BERT-large.
Successful predictions demonstrate early alignment of the subject (\textit{customers}; highlighted in red) with the verb (\textit{prefer}) around layer 5, facilitating accurate subject-verb agreement.
Conversely, in unsuccessful cases, this alignment emerged considerably later (around layer 12), highlighting delayed macro-syntactic integration.\looseness-1

These analyses suggest an optimal intermediate range of layers for integrating macro-syntactic information.
Forming macro-syntactic structures either prematurely or excessively late can negatively affect syntactic processing, highlighting the importance of appropriately timed integration for accurate predictions.
These visualizations underscore how deviations from this optimal timing contribute to subject-verb agreement errors.

\section{Discussion and Conclusion}
In this paper, we introduced \emph{Derivational Probing}---a method that integrates structural probing with an expected layer metric to trace the construction process of syntactic structures in neural language models. Our experiments revealed that BERT models tend to build micro-syntactic dependencies first and gradually assemble them into a coherent macro-structure.

BERT’s bidirectional context supports a stepwise, bottom-up construction---starting with the formation of local, micro-syntactic structures and culminating in a fully integrated macro representation. These findings offer valuable insights into the internal mechanisms by which deep neural models construct syntactic trees and highlight the importance of examining layer-wise structural formation for improved model interpretability.

One promising direction for future research is to incorporate multilingual probes, which will help determine whether these syntactic structures generalize beyond English or are not mere artifacts of the particular training corpus.
Another exciting direction would be to explore incremental parsing strategies in autoregressive language models as an alternative to non-incremental derivation processes across layers~\citep[cf.][]{eisape-etal-2022-probing}, which could yield further insights into the syntactic knowledge of neural language models.

\section*{Limitations}
First, our experiments were conducted on only two neural language models (BERT-base and BERT-large). It remains unclear whether similar results would be obtained for larger models or other architectural variants.
However, our method is applicable to any open neural model, making it feasible to extend this analysis to a broader range of models in future research.

Second, this study focused solely on English data.
It is uncertain whether similar layer-wise syntactic structure construction patterns would be observed when applying our method to other languages.
Nevertheless, our approach is language-agnostic, making cross-linguistic analysis an important direction for future work.

Furthermore, semantic cues may influence the results of syntactic probes. Our study does not fully account for these potential semantic confounds.
Future research should consider methods to more rigorously isolate syntactic information, such as using Jabberwocky sentences as demonstrated by \citet{hall-maudslay-cotterell-2021-syntactic}.

Lastly, our method relies on dependency parsing, primarily due to the use of the structural probe from \citet{hewitt2019structural}, which analyzes distances between tokens in the embedding space. This approach is inherently tied to formalisms like dependency grammar that focus on relationships between terminal symbols (tokens).
As a result, our method may not be directly applicable to other grammatical theories or parsing approaches that involve non-terminal symbols, such as constituency grammar.
This limitation arises because analyzing distances between tokens does not capture the hierarchical structures represented by non-terminals.
Future work could explore adapting our method or developing new probing techniques that can handle non-terminal representations to verify the generalizability of our findings.

\section*{Ethical considerations}
The training corpus is extracted from public web pages and thus could be socially biased, despite its popular use in the NLP community.

\section*{Acknowledgments}
This work was supported by JSPS KAKENHI Grant Numbers JP24H00087, JP24H00809 and JST PRESTO Grant Numbers JPMJPR21C2, JPMJPR21C8.
TS was also supported by JST BOOST NAIS.

\bibliography{custom}

\newpage
\appendix

\begin{figure*}[ht]
    \centering
    \includegraphics[width=\linewidth]{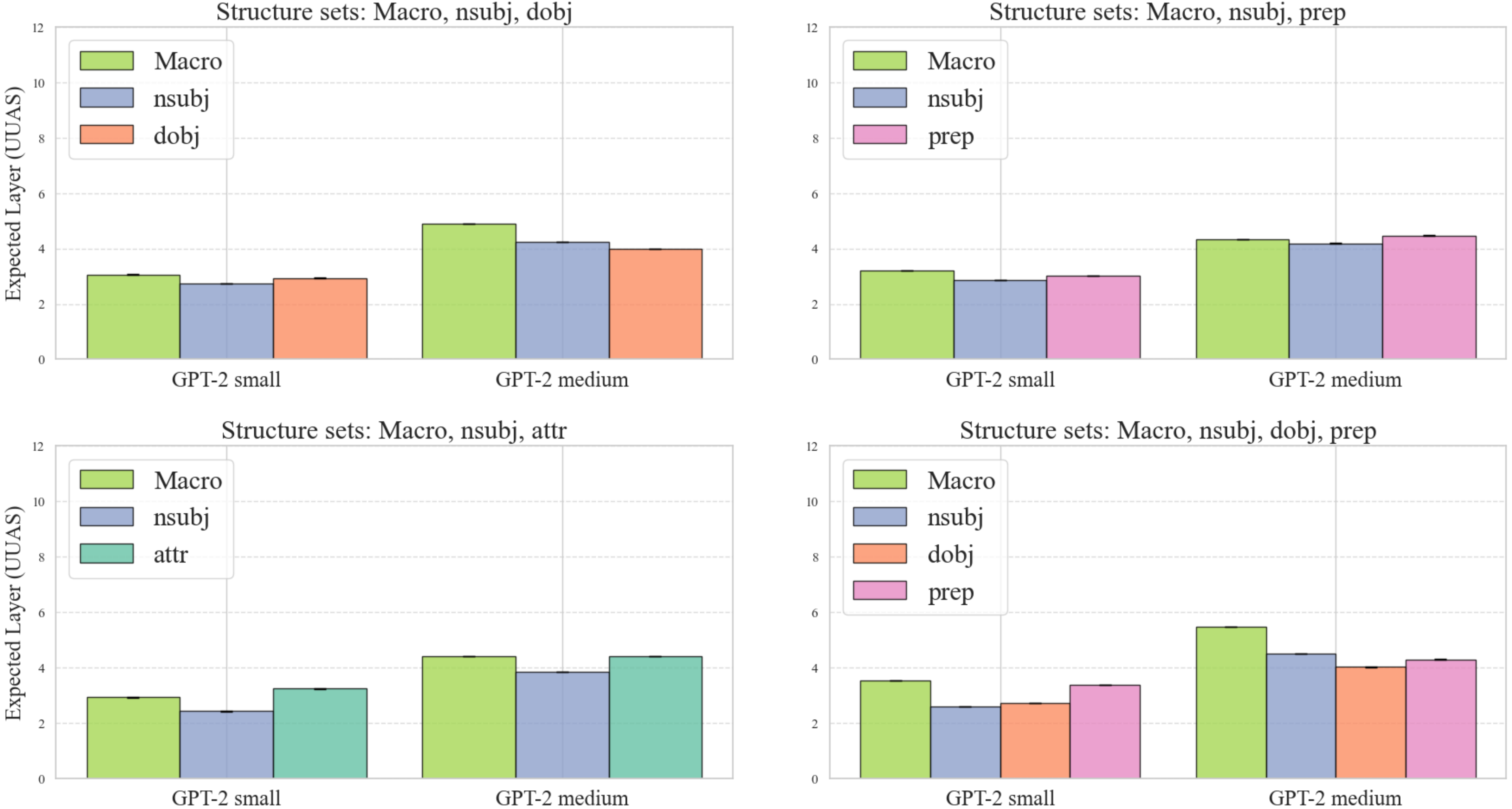}
    \caption{Expected layer for each GPT-2 model across different structure sets. Error bars represent standard deviation across 5 random seeds.}
    \label{fig:result_by_structure_sets_gpt2}
\end{figure*}

\begin{figure}[ht]
    \centering
    \includegraphics[width=\linewidth]{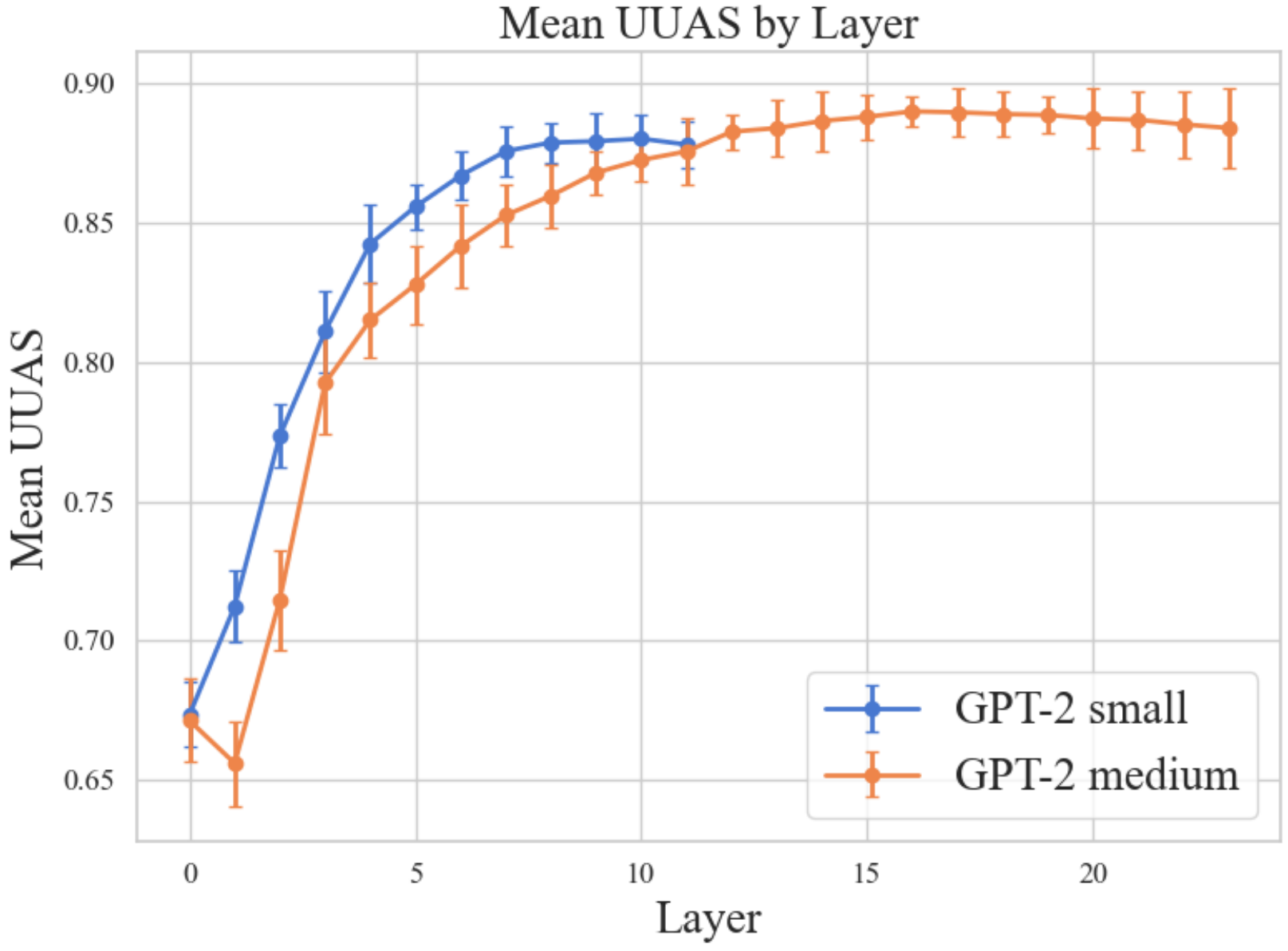}
    \caption{Global UUAS by each layer for each GPT-2 model. Error bars represent standard deviations across 5 random seeds.}
    \label{fig:global_uuas_gpt2}
\end{figure}

\section{The Experimental Results for GPT-2 Models} \label{sec:gpt2_result}
\Cref{fig:global_uuas_gpt2,fig:result_by_structure_sets_gpt2} show the experimental results with the same experimental setup as \Cref{sec:experimental_setup}, but conducted with GPT-2 small\footnote{\url{https://huggingface.co/openai-community/gpt2}}/medium\footnote{\url{https://huggingface.co/openai-community/gpt2-medium}}~\citep{radford2019language}.
In contrast to BERT, GPT-2 (both small and medium) displays a more \emph{parallel} derivation: the expected layer values for both micro-syntactic and macro-syntactic structures are closely aligned, suggesting parallel derivation rather than a bottom-up or top-down derivation.

\section{Hyperparameters} \label{sec:hparams}
Hyperparameters for our experiments are shown in \Cref{tab:params}. All models were trained and evaluated on 4×
NVIDIA RTX A5000 (24GB). The total computational cost for all experiments in this paper is about 120 GPU hours.

\begin{table}[h]
    \centering
    \begin{tabular}{lc}
        \toprule
         Optimizer & Adam \\
         Learning rate & 1e-3 \\
         Number of epochs & 40 \\
         Learning rate scheduler & ReduceLROnPlateau \\
         Batch size & 32 \\
         \toprule
    \end{tabular}
    \caption{Hyperparameters for our experiments}
    \label{tab:params}
\end{table}

\section{License of the Data and Tools}
The licenses of the data and tools used in this paper are summarized in \Cref{tab:license}. We confirmed that all the data and the tools were used under their respective license terms.
\begin{table}[h]
    \small
    \centering
    \begin{tabular}{p{5cm}p{2cm}}
    \toprule
    Data/tool     &  License\\
    \midrule
    \texttt{spacy}~\citep{Honnibal_spaCy_Industrial-strength_Natural_2020}& MIT\\
    \texttt{transformers}~\citep{wolf-etal-2020-transformers} & Apache 2.0\\
    \texttt{Wikitext-103}~\citep{merity2016pointer} & CC-BY-SA 3.0\\
    \bottomrule
    \end{tabular}
    \caption{License of the data and tools}
    \label{tab:license}
\end{table}

\end{document}

%% file: 00_packages.tex
\usepackage{times}
\usepackage{latexsym}
\usepackage{amsmath,bm}
\usepackage{amssymb}
\usepackage{hyperref}
\usepackage{cleveref}
\usepackage{booktabs}
\usepackage{linguex}
\usepackage{fontawesome}
\usepackage{tikz}
\usepackage{tikz-dependency}
\usepackage{xcolor}

%% file: 01_commands.tex
\definecolor{MacroColor}{RGB}{0,120,148}
\ifanonymoussubmissionmode
    
\else
    \ifarxivmode
        
    \else
        
    \fi
\fi

\crefname{section}{\S}{\S\S} 
\Crefname{section}{\S}{\S\S} 
\crefname{lemma}{Lemma}{Lemmata}
\crefname{table}{Table}{Tables}
\crefname{figure}{Figure}{Figures}
\crefname{algorithm}{Algorithm}{}
\crefname{equation}{Eq.}{Eqs.}
\crefname{appendix}{App.}{Appendices}
\crefformat{section}{\S#2#1#3}  

\newcommand{\depMacro}{\texttt{Marco}}
\newcommand{\depnsubj}{\texttt{nsubj}}
\newcommand{\depdobj}{\texttt{dobj}}
\newcommand{\depprep}{\texttt{prep}}
\newcommand{\depattr}{\texttt{attr}}

\newcommand{\softmax}{\mathrm{softmax}}
\newcommand{\mixingweight}{\mathbf{w}}

%% file: main.bbl
\begin{thebibliography}{24}
\providecommand{\natexlab}[1]{#1}

\bibitem[{Chang and Bergen(2024)}]{10.1162/coli_a_00492}
Tyler~A. Chang and Benjamin~K. Bergen. 2024.
\newblock \href {https://doi.org/10.1162/coli_a_00492} {{Language Model Behavior: A Comprehensive Survey}}.
\newblock \emph{Computational Linguistics}, 50(1):293--350.

\bibitem[{Clark et~al.(2019)Clark, Khandelwal, Levy, and Manning}]{clark-etal-2019-bert}
Kevin Clark, Urvashi Khandelwal, Omer Levy, and Christopher~D. Manning. 2019.
\newblock \href {https://doi.org/10.18653/v1/W19-4828} {What does {BERT} look at? an analysis of {BERT}`s attention}.
\newblock In \emph{Proceedings of the 2019 ACL Workshop BlackboxNLP: Analyzing and Interpreting Neural Networks for NLP}, pages 276--286, Florence, Italy. Association for Computational Linguistics.

\bibitem[{Devlin et~al.(2019)Devlin, Chang, Lee, and Toutanova}]{devlin-etal-2019-bert}
Jacob Devlin, Ming-Wei Chang, Kenton Lee, and Kristina Toutanova. 2019.
\newblock \href {https://doi.org/10.18653/v1/N19-1423} {{BERT}: Pre-training of deep bidirectional transformers for language understanding}.
\newblock In \emph{Proceedings of the 2019 Conference of the North {A}merican Chapter of the Association for Computational Linguistics: Human Language Technologies, Volume 1 (Long and Short Papers)}, pages 4171--4186, Minneapolis, Minnesota. Association for Computational Linguistics.

\bibitem[{Eisape et~al.(2022)Eisape, Gangireddy, Levy, and Kim}]{eisape-etal-2022-probing}
Tiwalayo Eisape, Vineet Gangireddy, Roger Levy, and Yoon Kim. 2022.
\newblock \href {https://doi.org/10.18653/v1/2022.findings-emnlp.203} {Probing for incremental parse states in autoregressive language models}.
\newblock In \emph{Findings of the Association for Computational Linguistics: EMNLP 2022}, pages 2801--2813, Abu Dhabi, United Arab Emirates. Association for Computational Linguistics.

\bibitem[{Hayashi et~al.(2012)Hayashi, Watanabe, Asahara, and Matsumoto}]{hayashi-etal-2012-head}
Katsuhiko Hayashi, Taro Watanabe, Masayuki Asahara, and Yuji Matsumoto. 2012.
\newblock \href {https://aclanthology.org/P12-1069/} {Head-driven transition-based parsing with top-down prediction}.
\newblock In \emph{Proceedings of the 50th Annual Meeting of the Association for Computational Linguistics (Volume 1: Long Papers)}, pages 657--665, Jeju Island, Korea. Association for Computational Linguistics.

\bibitem[{Hewitt et~al.(2021)Hewitt, Ethayarajh, Liang, and Manning}]{hewitt-etal-2021-conditional}
John Hewitt, Kawin Ethayarajh, Percy Liang, and Christopher Manning. 2021.
\newblock \href {https://doi.org/10.18653/v1/2021.emnlp-main.122} {Conditional probing: measuring usable information beyond a baseline}.
\newblock In \emph{Proceedings of the 2021 Conference on Empirical Methods in Natural Language Processing}, pages 1626--1639, Online and Punta Cana, Dominican Republic. Association for Computational Linguistics.

\bibitem[{Hewitt and Manning(2019)}]{hewitt2019structural}
John Hewitt and Christopher~D. Manning. 2019.
\newblock \href {https://doi.org/10.18653/v1/N19-1419} {{A} structural probe for finding syntax in word representations}.
\newblock In \emph{Proceedings of the 2019 Conference of the North {A}merican Chapter of the Association for Computational Linguistics: Human Language Technologies, Volume 1 (Long and Short Papers)}, pages 4129--4138, Minneapolis, Minnesota. Association for Computational Linguistics.

\bibitem[{Honnibal et~al.(2020)Honnibal, Montani, Van~Landeghem, and Boyd}]{Honnibal_spaCy_Industrial-strength_Natural_2020}
Matthew Honnibal, Ines Montani, Sofie Van~Landeghem, and Adriane Boyd. 2020.
\newblock \href {https://doi.org/10.5281/zenodo.1212303} {{spaCy: Industrial-strength Natural Language Processing in Python}}.

\bibitem[{Kunz and Kuhlmann(2022)}]{kunz-kuhlmann-2022-linguistic}
Jenny Kunz and Marco Kuhlmann. 2022.
\newblock \href {https://aclanthology.org/2022.coling-1.413/} {Where does linguistic information emerge in neural language models? measuring gains and contributions across layers}.
\newblock In \emph{Proceedings of the 29th International Conference on Computational Linguistics}, pages 4664--4676, Gyeongju, Republic of Korea. International Committee on Computational Linguistics.

\bibitem[{Limisiewicz and Mare{\v{c}}ek(2021)}]{limisiewicz-marecek-2021-introducing}
Tomasz Limisiewicz and David Mare{\v{c}}ek. 2021.
\newblock \href {https://doi.org/10.18653/v1/2021.acl-long.36} {Introducing orthogonal constraint in structural probes}.
\newblock In \emph{Proceedings of the 59th Annual Meeting of the Association for Computational Linguistics and the 11th International Joint Conference on Natural Language Processing (Volume 1: Long Papers)}, pages 428--442, Online. Association for Computational Linguistics.

\bibitem[{Marvin and Linzen(2018)}]{marvin-linzen-2018-targeted}
Rebecca Marvin and Tal Linzen. 2018.
\newblock \href {https://doi.org/10.18653/v1/D18-1151} {Targeted syntactic evaluation of language models}.
\newblock In \emph{Proceedings of the 2018 Conference on Empirical Methods in Natural Language Processing}, pages 1192--1202, Brussels, Belgium. Association for Computational Linguistics.

\bibitem[{Maudslay and Cotterell(2021)}]{hall-maudslay-cotterell-2021-syntactic}
Rowan~Hall Maudslay and Ryan Cotterell. 2021.
\newblock \href {https://doi.org/10.18653/v1/2021.naacl-main.11} {Do syntactic probes probe syntax? experiments with jabberwocky probing}.
\newblock In \emph{Proceedings of the 2021 Conference of the North American Chapter of the Association for Computational Linguistics: Human Language Technologies}, pages 124--131, Online. Association for Computational Linguistics.

\bibitem[{Merity et~al.(2016)Merity, Xiong, Bradbury, and Socher}]{merity2016pointer}
Stephen Merity, Caiming Xiong, James Bradbury, and Richard Socher. 2016.
\newblock \href {https://arxiv.org/abs/1609.07843} {Pointer sentinel mixture models}.
\newblock \emph{Preprint}, arXiv:1609.07843.

\bibitem[{Nivre(2004)}]{nivre-2004-incrementality}
Joakim Nivre. 2004.
\newblock \href {https://aclanthology.org/W04-0308/} {Incrementality in deterministic dependency parsing}.
\newblock In \emph{Proceedings of the Workshop on Incremental Parsing: Bringing Engineering and Cognition Together}, pages 50--57, Barcelona, Spain. Association for Computational Linguistics.

\bibitem[{Pedregosa et~al.(2011)Pedregosa, Varoquaux, Gramfort, Michel, Thirion, Grisel, Blondel, Prettenhofer, Weiss, Dubourg, Vanderplas, Passos, Cournapeau, Brucher, Perrot, and {{\'E}}douard Duchesnay}]{scikit-learn}
Fabian Pedregosa, Ga{{\"e}}l Varoquaux, Alexandre Gramfort, Vincent Michel, Bertrand Thirion, Olivier Grisel, Mathieu Blondel, Peter Prettenhofer, Ron Weiss, Vincent Dubourg, Jake Vanderplas, Alexandre Passos, David Cournapeau, Matthieu Brucher, Matthieu Perrot, and {{\'E}}douard Duchesnay. 2011.
\newblock \href {http://jmlr.org/papers/v12/pedregosa11a.html} {Scikit-learn: Machine learning in python}.
\newblock \emph{Journal of Machine Learning Research}, 12(85):2825--2830.

\bibitem[{Peters et~al.(2018)Peters, Neumann, Iyyer, Gardner, Clark, Lee, and Zettlemoyer}]{peters-etal-2018-deep}
Matthew~E. Peters, Mark Neumann, Mohit Iyyer, Matt Gardner, Christopher Clark, Kenton Lee, and Luke Zettlemoyer. 2018.
\newblock \href {https://doi.org/10.18653/v1/N18-1202} {Deep contextualized word representations}.
\newblock In \emph{Proceedings of the 2018 Conference of the North {A}merican Chapter of the Association for Computational Linguistics: Human Language Technologies, Volume 1 (Long Papers)}, pages 2227--2237, New Orleans, Louisiana. Association for Computational Linguistics.

\bibitem[{Prim(1957)}]{prim_1957}
R.~C. Prim. 1957.
\newblock \href {https://doi.org/10.1002/j.1538-7305.1957.tb01515.x} {Shortest connection networks and some generalizations}.
\newblock \emph{Bell System Technical Journal}, 36(6):1389--1401.

\bibitem[{Radford et~al.(2019)Radford, Wu, Child, Luan, Amodei, and Sutskever}]{radford2019language}
Alec Radford, Jeff Wu, Rewon Child, David Luan, Dario Amodei, and Ilya Sutskever. 2019.
\newblock Language models are unsupervised multitask learners.

\bibitem[{Salazar et~al.(2020)Salazar, Liang, Nguyen, and Kirchhoff}]{salazar-etal-2020-masked}
Julian Salazar, Davis Liang, Toan~Q. Nguyen, and Katrin Kirchhoff. 2020.
\newblock \href {https://doi.org/10.18653/v1/2020.acl-main.240} {Masked language model scoring}.
\newblock In \emph{Proceedings of the 58th Annual Meeting of the Association for Computational Linguistics}, pages 2699--2712, Online. Association for Computational Linguistics.

\bibitem[{Tenney et~al.(2019)Tenney, Das, and Pavlick}]{tenney2019bert}
Ian Tenney, Dipanjan Das, and Ellie Pavlick. 2019.
\newblock \href {https://doi.org/10.18653/v1/P19-1452} {{BERT} rediscovers the classical {NLP} pipeline}.
\newblock In \emph{Proceedings of the 57th Annual Meeting of the Association for Computational Linguistics}, pages 4593--4601, Florence, Italy. Association for Computational Linguistics.

\bibitem[{Vig and Belinkov(2019)}]{vig-belinkov-2019-analyzing}
Jesse Vig and Yonatan Belinkov. 2019.
\newblock \href {https://doi.org/10.18653/v1/W19-4808} {Analyzing the structure of attention in a transformer language model}.
\newblock In \emph{Proceedings of the 2019 ACL Workshop BlackboxNLP: Analyzing and Interpreting Neural Networks for NLP}, pages 63--76, Florence, Italy. Association for Computational Linguistics.

\bibitem[{White et~al.(2021)White, Pimentel, Saphra, and Cotterell}]{white-etal-2021-non}
Jennifer~C. White, Tiago Pimentel, Naomi Saphra, and Ryan Cotterell. 2021.
\newblock \href {https://doi.org/10.18653/v1/2021.naacl-main.12} {A non-linear structural probe}.
\newblock In \emph{Proceedings of the 2021 Conference of the North American Chapter of the Association for Computational Linguistics: Human Language Technologies}, pages 132--138, Online. Association for Computational Linguistics.

\bibitem[{Wolf et~al.(2020)Wolf, Debut, Sanh, Chaumond, Delangue, Moi, Cistac, Rault, Louf, Funtowicz, Davison, Shleifer, von Platen, Ma, Jernite, Plu, Xu, Le~Scao, Gugger, Drame, Lhoest, and Rush}]{wolf-etal-2020-transformers}
Thomas Wolf, Lysandre Debut, Victor Sanh, Julien Chaumond, Clement Delangue, Anthony Moi, Pierric Cistac, Tim Rault, Remi Louf, Morgan Funtowicz, Joe Davison, Sam Shleifer, Patrick von Platen, Clara Ma, Yacine Jernite, Julien Plu, Canwen Xu, Teven Le~Scao, Sylvain Gugger, Mariama Drame, Quentin Lhoest, and Alexander Rush. 2020.
\newblock \href {https://doi.org/10.18653/v1/2020.emnlp-demos.6} {Transformers: State-of-the-art natural language processing}.
\newblock In \emph{Proceedings of the 2020 Conference on Empirical Methods in Natural Language Processing: System Demonstrations}, pages 38--45, Online. Association for Computational Linguistics.

\bibitem[{Zhao et~al.(2024)Zhao, Chen, Yang, Liu, Deng, Cai, Wang, Yin, and Du}]{10.1145/3639372}
Haiyan Zhao, Hanjie Chen, Fan Yang, Ninghao Liu, Huiqi Deng, Hengyi Cai, Shuaiqiang Wang, Dawei Yin, and Mengnan Du. 2024.
\newblock \href {https://doi.org/10.1145/3639372} {Explainability for large language models: A survey}.
\newblock \emph{ACM Trans. Intell. Syst. Technol.}, 15(2).

\end{thebibliography}
